\pdfoutput=1
\documentclass[letterpaper, 10 pt, conference]{ieeeconf}
\IEEEoverridecommandlockouts 

\usepackage[utf8]{inputenc}
\usepackage[scaled=0.86]{helvet}

\usepackage{amsmath,amssymb,amsfonts, mathtools, amsthm, nccmath}
\usepackage{marvosym}
\usepackage{graphicx}
\usepackage{floatrow}
\usepackage{xcolor}
\usepackage[binary-units=true, per-mode=symbol]{siunitx}
\usepackage{booktabs}
\usepackage{nicefrac}
\usepackage[scaled=.8]{beramono}
\usepackage{ifthen}
\usepackage{censor}
\usepackage{svg}
\usepackage{floatrow}
\usepackage{mdframed}
\usepackage{dblfloatfix}
\usepackage[bottom]{footmisc}
\usepackage{pict2e}
\usepackage{import}
\usepackage{subcaption}
\usepackage{xcolor}
\usepackage{bm}
\usepackage[normalem]{ulem}
\newcommand{\vect}[1]{\boldsymbol{\mathbf{#1}}}

\usepackage{booktabs}
\usepackage{textcomp}
\usepackage{color,soul}
\usepackage[font={small}, figurename={Fig.}, tablename={Tab.}]{caption}
\usepackage{xspace}
\usepackage{balance}
\usepackage{mathtools}
\usepackage{tabularx}
\usepackage{multirow}
\usepackage{listings}
\usepackage{booktabs}
\usepackage{float}
\floatstyle{plaintop}
\restylefloat{table}
\usepackage{etoolbox}
\usepackage{nicefrac} 

\renewcommand\hl[1]{#1} 

\usepackage{tikz}
\usetikzlibrary{decorations.pathreplacing,calc}
\newcommand{\tikzmark}[1]{\tikz[overlay,remember picture] \node (#1) {};}

\newcommand*{\SmallMargin}{0.5cm}%

\newcommand*{\AddNote}[4]{%
    \begin{tikzpicture}[overlay, remember picture]
        \draw [decoration={brace,amplitude=0.5em},decorate,thick,black]
            ($(#3)!(#1.north)!($(#3)-(0,1)$)$) --  
            ($(#3)!(#2.south)!($(#3)-(0,1)$)$)
                node [align=center, text width=2.5cm, pos=0.5, anchor=west] {#4};
    \end{tikzpicture}
}%

\makeatletter
\newcommand{\labeltext}[2]{%
  \@bsphack
  \csname phantomsection\endcsname 
  \def\@currentlabel{#1}{\label{#2}}%
  \@esphack
}
\makeatother

\usepackage{bbm}
\usepackage{booktabs,siunitx}

\usepackage[percent]{overpic}
\usepackage{siunitx}
\usepackage[style=ieee, backend=biber, bibencoding=utf8, maxbibnames=10, mincitenames=1, maxcitenames=2, url=false, doi=false, isbn=false, citestyle=ieee, natbib=true, eprint=true, alldates=year]{biblatex}
\DeclareSourcemap{
  \maps[datatype=bibtex, overwrite]{
    \map{
        \step[fieldset=address, null]
        \step[fieldset=location, null]
        \step[fieldset=editor, null]
        \step[fieldset=series, null]
        \step[fieldsource=issue, match=\regexp{\A[0-9]+\Z}, final]
        \step[fieldset=number, origfieldval]
        \step[fieldset=issue, null]
    }
    \map{
        \pertype{inproceedings}
        \step[fieldset=publisher, null]
        \step[fieldset=volume, null]
    }
  }
}

\usepackage[hidelinks]{hyperref}
\usepackage[utf8]{inputenc}
\usepackage[T1]{fontenc}
\usepackage{listings}
\usepackage{xcolor}

\usepackage{algorithm}
\usepackage{algorithmicx}
\usepackage[noend]{algpseudocode}

\algnewcommand\algorithmicconfig{\textbf{Configuration:}}
\algnewcommand\Configuration{\item[\algorithmicconfig]}
\algnewcommand\algorithmicinit{\textbf{Initialization:}}
\algnewcommand\Initialization{\item[\algorithmicinit]}
\algnewcommand\algorithmicproc{\textbf{Procedure:}}
\algnewcommand\Proc{\item[\algorithmicproc]}

\makeatletter
\newcommand\setalgorithmcaptionfont[1]{%
  \let\my@floatc@ruled\floatc@ruled          
  \def\floatc@ruled{%
    \global\let\floatc@ruled\my@floatc@ruled 
    #1\floatc@ruled}}
\makeatother
 
\definecolor{conflict}{rgb}{0.97,0.15,0.52}
\definecolor{codegreen}{rgb}{0,0.6,0}
\definecolor{codegray}{rgb}{0.5,0.5,0.5}
\definecolor{codepurple}{rgb}{0.58,0,0.82}
\definecolor{backcolour}{rgb}{0.95,0.95,0.92}
 
\lstdefinestyle{mystyle}{
    backgroundcolor=\color{backcolour},   
    commentstyle=\color{codegreen},
    keywordstyle=\color{magenta},
    numberstyle=\tiny\color{codegray},
    stringstyle=\color{codepurple},
    basicstyle=\footnotesize,
    breakatwhitespace=false,         
    breaklines=true,                 
    captionpos=b,                    
    keepspaces=true,                 
    numbers=left,                    
    numbersep=5pt,                  
    showspaces=false,                
    showstringspaces=false,
    showtabs=false,                  
    tabsize=2
}

\newcommand\fauxsc[1]{\fauxschelper#1 \relax\relax}
\def\fauxschelper#1 #2\relax{%
  \fauxschelphelp#1\relax\relax%
  \if\relax#2\relax\else\ \fauxschelper#2\relax\fi%
}
\def\Hscale{.85}\def\Vscale{.74}\def\Cscale{1.12}
\def\fauxschelphelp#1#2\relax{%
  \ifnum`#1>``\ifnum`#1<`\{\scalebox{\Hscale}[\Vscale]{\uppercase{#1}}\else%
    \scalebox{\Cscale}[1]{#1}\fi\else\scalebox{\Cscale}[1]{#1}\fi%
  \ifx\relax#2\relax\else\fauxschelphelp#2\relax\fi}

\usepackage[capitalise, nameinlink]{cleveref}
\crefformat{equation}{(#2#1#3)}

\DeclareMathOperator{\atantwo}{atan2}

\usepackage{footnote}
\usepackage{tablefootnote}
\makesavenoteenv{tabular}
\makesavenoteenv{table}
\usepackage{longtable}
\usepackage{xhfill}
\newcommand{\ditto}[1][.4pt]{\xrfill{#1}~\textquotedbl~\xrfill{#1}}
\PassOptionsToPackage{hyphens}{url}
\urldef{\heterodataurl}\url{https://pytorch-geometric.readthedocs.io/en/latest/modules/data.html#torch_geometric.data.HeteroData}
\Crefname{figure}{Fig.}{Figs.}
\Crefname{table}{Tab.}{Tabs.}
\Crefname{algorithm}{Alg.}{Algs.}

\newcommand{\packagename}{{\textit{cr-geo}}}
\newcommand{\pygeo}{\textit{PyG}}

\definecolor{britishracinggreen}{rgb}{0.0, 0.7, 0.1}
\definecolor{darkblue}{rgb}{0.0, 0.15, 0.5}

\DeclareRobustCommand{\hll}[1]{{\sethlcolor{cyan}\hl{#1}}}
\DeclareRobustCommand{\hlll}[1]{{\sethlcolor{green}\hl{#1}}}
\DeclareRobustCommand{\hllll}[1]{{\sethlcolor{pink}\hl{#1}}}

\newcommand{\para}[1]{{}}

\usepackage{xcolor}

\usepackage{color}
\definecolor{tumblue}{rgb}{0, 0.4, 0.74}
\definecolor{darkgreen}{rgb}{0.0, 0.5, 0.0}
\definecolor{lightgreen}{rgb}{0.0, 0.84, 0.0}

\def\code#1{\textit{#1}}
\def\ccode#1{\texttt{\normalsize{#1}}}

\newcommand{\NA}{---}

\makeatletter
\newcommand{\uset}[2]{%
  {\mathop{#2}\limits^{\vbox to -.5\ex@{\kern-\tw@\ex@
   \hbox{\scriptsize #1}\vss}}}}
\makeatother

\usepackage{siunitx}
\usepackage[acronym, nohypertypes={acronym}]{glossaries}

\newcolumntype{b}{X}
\newcolumntype{s}{>{\hsize=.7\hsize}X}
\newcolumntype{w}{>{\hsize=.3\hsize}X}
\newcolumntype{t}{>{\hsize=.45\hsize}X}
\newcolumntype{W}{>{\hsize=.2\hsize}X}
\newcolumntype{E}{>{\hsize=.1\hsize}X}
\newcolumntype{R}{>{\hsize=.05\hsize}X}

\makeatletter
    \algrenewcommand\alglinenumber[1]{\tikzmark{\arabic{ALG@line}}}
\makeatother

\makeatletter
\newcounter{phase}[algorithm]
\newlength{\phaserulewidth}

\makeatother



\newcolumntype{L}[1]{>{\raggedright\arraybackslash}p{#1}}

\newcolumntype{C}[1]{>{\centering\arraybackslash}p{#1}}

\newcolumntype{R}[1]{>{\raggedleft\arraybackslash}p{#1}}

\title{\LARGE \bf \hl{Geometric Deep Learning for Autonomous Driving: Unlocking the Power of Graph Neural Networks With CommonRoad-Geometric}}

\author{Eivind Meyer, Maurice Brenner, Bowen Zhang, Max Schickert, Bilal Musani, and Matthias Althoff
    \thanks{Department of Informatics, Technical University of Munich, Garching, Germany. \url{{eivind.meyer, maurice.brenner, bowen.zhang, max.schickert, bilal.musani, althoff}@tum.de}}%
}

\begin{document}

\lstset{language=Python}

\newtheoremstyle{bfnote}%
  {}{}
  {\itshape}{}
  {\bfseries}{.}
  { }{\thmname{#1}\thmnote{ (#3)}}
\theoremstyle{bfnote}
\newtheorem{assumption}{Assumption}
\newtheorem{definition}{Definition}
\newtheorem{remark}{\small Remark}

\maketitle
\thispagestyle{plain}
\pagestyle{plain}

\begin{abstract}
Heterogeneous graphs offer powerful data representations \hll{for traffic, given their ability to model the complex interaction} effects among a varying number of traffic participants and the underlying road infrastructure. With the recent advent of graph neural networks (GNNs) as the accompanying deep learning framework, \hll{the graph structure can be efficiently leveraged for various machine learning applications such as trajectory prediction.} As a first of its kind, our proposed \hlll{Python} framework offers an easy-to-use \hll{and} fully customizable data processing pipeline to extract standardized graph datasets from traffic scenarios. Providing a platform for \hll{GNN-based autonomous driving} research, it improves comparability between approaches and allows researchers to focus on model implementation instead of dataset curation.
\end{abstract}



\section{Introduction}

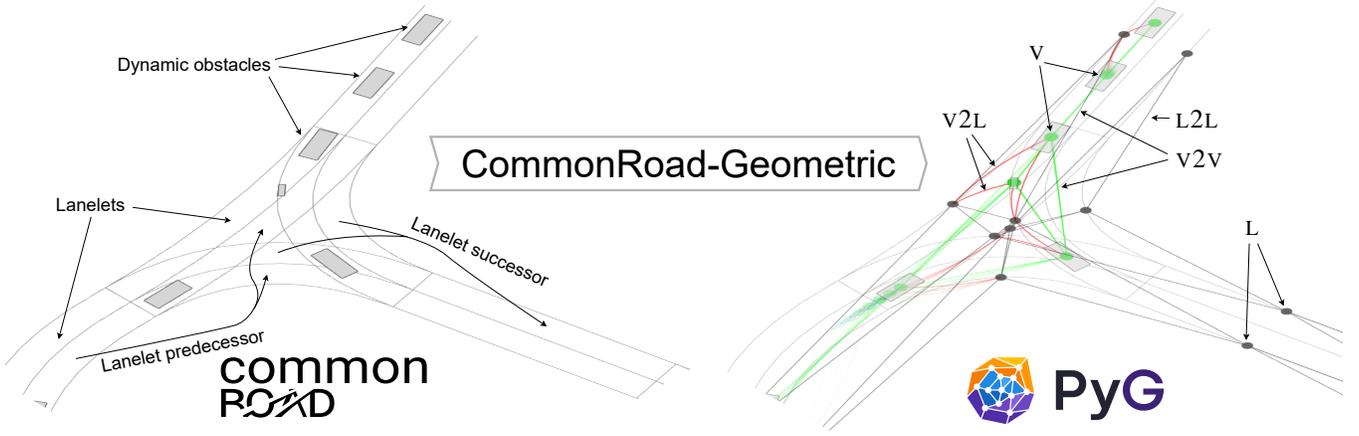
\begin{figure*}[t]
    \centering

    \begin{overpic}[trim=400 265 500 100, clip, width=1.0\linewidth]{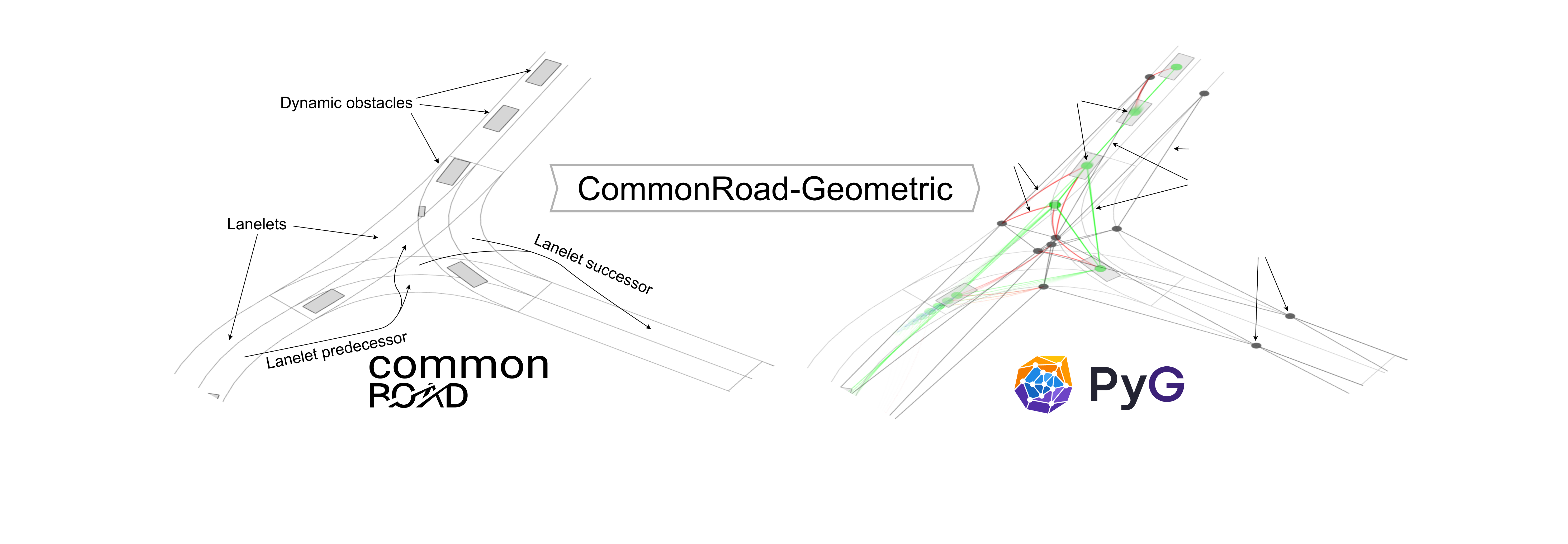}
    \caption{Our package serves as a bridge from CommonRoad to PyTorch-Geometric.
    \hl{The abbreviated graph labels refer to the heterogeneous node and edge entities of our unified traffic graph structure covered in}~\cref{section:graph-representation}.
    } 
 \put (70, 23) {\textsc{v2l}}
 \put (87.5, 20) {\textsc{v2v}}
 \put (87.5, 23) {\textsc{l2l}}
 \put (76.5, 28) {\textsc{v}}
 \put (92.7,14.9) {\textsc{l}}
 \label{fig:crgeo_highlevel}
   \end{overpic}
\end{figure*}

\hl{Machine learning agents require an accurate understanding of the surrounding traffic context to make safe and effective decisions}~\cite{Lefevre2014}.
This calls for a descriptive representation not only of the various entities within the traffic environment, but also 
their complex spatial and temporal relationships. When restricted to a fixed-size feature space\hll{---a prerequisite for classical neural networks}---this is a particularly challenging prospect given the inherent complexity and variability in road geometries and traffic situations.

An alternative approach is to model the environment state in the form of a heterogeneous graph, encompassing both the road network topology and the traffic participants present in it. \hl{This structured representation} \hll{allows us to capture} \hl{a wide range of road networks and traffic scenarios with a variable number of discrete elements. Additionally, it enables the explicit modeling of pairwise relationships or interactions between entities through edge connections.} \hll{Graph neural networks (GNNs) have recently emerged as the principal deep learning framework for processing graph data}~\cite{bronstein_2017, chami_machine_2021, hetero_gnn_wang_2019, battaglia_2018}\hll{. By using GNNs, the graph topology can be exploited as a relational inductive  bias}~\cite{battaglia_2018}\hll{ during training, aiding generalization.}

\subsection{Related work}

\hll{Graph-structured} representations of traffic \hll{proposed in existing works} differ based on the objectives of the \hl{learning task}. In~\cite{jepsen_2019}, the authors let nodes represent individual road segments with edges forming the overarching road network topology \hl{for driving speed estimation and road network classification.} Alternatively,~\cite{diehl_2019} and~\cite{jeon2020_scalenet} 
present graph-based traffic forecasting approaches in which nodes are used to model traffic participants and edges are used to capture vehicle interactions. Similarly, \autocite{huegle_dynamic_2019} and~\cite{hart_graph_2020} adopt vehicle-to-vehicle GNNs as policy networks for reinforcement learning agents.  \hl{Additionally considering the temporal dimension,} \autocite{liGRIPGraphbasedInteractionaware2019} employs a spatiotemporal vehicle graph for capturing time-dependent features. \hl{Finally, recent works} \cite{liang_lane_graph_2020, zeng_lanercnn_2021, zhao2020tnt, kim_lapred_2021, zhang2021trajectory, janjo2021starnet, ngiam2021scene, gilles2021gohome, mo_2022_edge_enhanced} \hl{incorporate both vehicle and map nodes by modelling} the environment as a heterogeneous graph, including both inter-vehicle as well as vehicle-road interaction \hl{effects}.
\subsection{Motivation and contributions}

Despite the vast research interest, there is no software framework that offers an interface for extracting custom graph datasets from traffic scenarios. \hl{Although there are plenty of autonomous driving datasets}, e.g. ~\cite{krajewskiHighDDatasetDrone2018, interactiondataset, caesar_nuplan_2022}\hl{, researchers have to write considerable amounts of ad-hoc, error-prone conversion code to use them as inputs for GNN models.} As evident by the success in other application domains such as bioinformatics and social networks~\cite{morris_2020_tudataset, leskovec_2024_snapsets}, standardized graph datasets would enable autonomous driving researchers to streamline their experiments and ensure comparability and repeatability of their results.

To fill this gap, we propose \textit{CommonRoad-Geometric} (\packagename): a Python library designed to facilitate the extraction of graph data from \hl{recorded or simulated} traffic scenarios. \hll{Our framework extends the} \textit{CommonRoad}~\cite{commonroad_paper} \hll{software platform} and uses its standardized interface for a high-level representation of the traffic environment.
\hll{As illustrated by}~\cref{fig:crgeo_highlevel}, our framework unifies \hl{the traffic scene} into a single graph entity, encompassing both the traffic participants and the underlying road map. The extracted graph representations are based on the \texttt{HeteroData} class offered by \code{PyTorch-Geometric}~\cite{pygeo}, a popular \code{PyTorch}~\cite{pytorch} \hl{extension} for deep learning on graph-structured data.

Our paper offers the following contributions:
\begin{itemize}
    \item \hll{we introduce a heterogeneous graph structure for map-aware traffic representations tailored to GNN applications;}
    \item \hll{we present and outline the software architecture of} \code{CommonRoad-Geometric} (\packagename)\hll{, which offers a bridge from the well-established CommonRoad scenario format to} \code{PyTorch-Geometric} (\pygeo);
    \item as a concrete example of the wide range of GNN-based applications facilitated by \packagename, we train a \hll{spatiotemporal trajectory prediction model on a real-world graph dataset extracted from NuPlan}~\cite{caesar_nuplan_2022}.
\end{itemize}

\footnotetext[1]{Our source code is available at {\footnotesize\url{https://github.com/CommonRoad/crgeo}}, and is provided under the BSD-3-Clause license, allowing free use and distribution.}

\section{Background}

We first provide some background on heterogeneous graphs and the description of scenarios in CommonRoad.

\subsubsection{Heterogeneous graphs}
\begin{figure}[htp]
    \centering
    \includegraphics[width=1.0\linewidth]{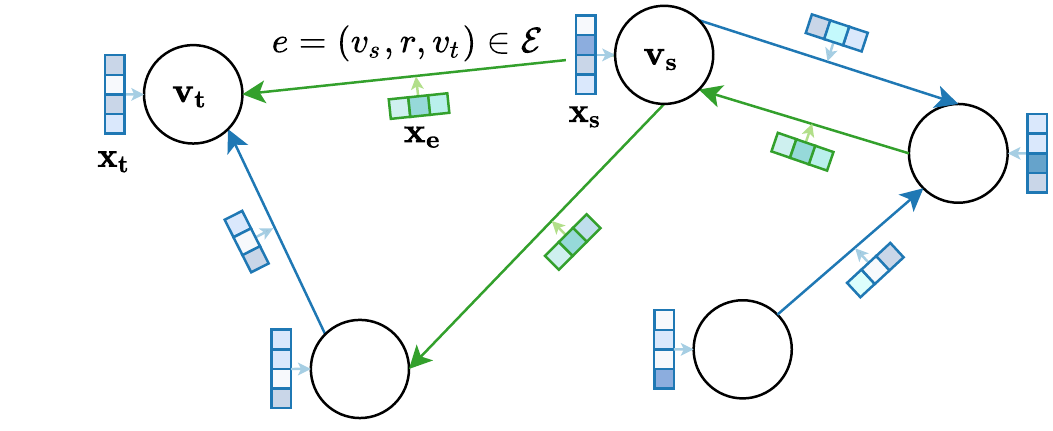}
    \caption{Illustration of a simple graph structure. Edge types are represented as different colors. Colored squares represent features vectors for nodes (e.g., vehicle speed) and edges (e.g., distance between vehicles).}
    \label{fig:simple_graph}
\end{figure}

A directed heterogeneous graph ${\mathcal{G} = (\mathcal{V}, \mathcal{E}, \mathcal{A}, \mathcal{R}, \mathcal{X}_{\mathcal{V}}, \mathcal{X}_{\mathcal{E}})}$ is defined as a tuple of a set of nodes $\mathcal{V}$, edges $\mathcal{E}$, node types $\mathcal{A}$, edge types $\mathcal{R}$, and corresponding node and edge features $\mathcal{X}_{\mathcal{V}}$ and $\mathcal{X}_{\mathcal{E}}$~\cite{jizhou_2013_heterograph}. Edges are defined as a 3-tuple ${\mathcal{E} \subset \mathcal{V} \times \mathcal{R} \times \mathcal{V}}$. Further, each node ${v \in \mathcal{V}}$ is assigned a node type via the mapping ${\tau_V(v): \mathcal{V} \rightarrow \mathcal{A}}$. Analogously, edges ${e \in \mathcal{E}}$ are associated with an edge type ${\tau_\mathcal{E}(e): \mathcal{E} \rightarrow \mathcal{R}}$. Finally, $\mathcal{X}_{\mathcal{V}}$ and $\mathcal{X}_{\mathcal{E}}$ contain the feature vectors, which we denote as ${\vect{x}_v \in \mathbb{R}^{D_\mathcal{V}}}$ for node features and ${\vect{x}_e \in \mathbb{R}^{D_\mathcal{E}}}$ for edge features. \cref{fig:simple_graph} illustrates graph features on the node and edge level.

\subsubsection{CommonRoad}

A CommonRoad \textit{scenario} contains a set $\mathcal{V}^{cr}$ of \textit{dynamic obstacles}. \hl{For a  vehicle $V \in \mathcal{V}^{cr}$ of rectangular shape $(l_V, w_V)$, we represent its time-dependent state by its x-y center position $\vect{p}_{V}$, its orientation $\theta_{V}$, as well as their time derivatives in the scenario coordinate frame.}

Further, the map-related information of a scenario is described by its \code{lanelet network}, containing a set $\mathcal{L}^{cr}$ of atomic \code{lanelets} and their \hllll{longitudinal and lateral}\hl{ adjacency relations~\cite{bender_lanelets}.}
Lanelets are geometrically defined by their boundary polylines. For a lanelet $L \in \mathcal{L}^{cr}$, we denote its left and right boundary polylines as $S_{L, l}$ and $S_{L, r}$, respectively. Further, we denote the center polyline as $S_{L, c}$. Each polyline is made up of a sequence of $N_{L}$ waypoints in the \hll{scenario} coordinate frame. Using the notation $\langle \square_{i} \rangle_{i \in 1, ..., N}$ to define a sequence of values of length $N$, we let the polylines be denoted as ${S_{L, l} = \langle \vec{s}_{l, i} \rangle_{i \in 1, ..., N_{L}}}$, ${S_{L, r} = \langle \vec{s}_{r, i} \rangle_{i \in 1, ..., N_{L}}}$, and ${S_{L, c} = \langle \vec{s}_{c, i} \rangle_{i \in 1, ..., N_{L}}}$.
With $\lVert \rVert_2$ denoting the L2-norm, we further define the lanelet length as
\begin{equation}
\begin{aligned}
\lVert L \rVert_2 &= \sum_{i=1}^{N_L - 1} \sqrt{\lVert \vec{s}_{c,i+1} - \vec{s}_{c,i} \rVert_2},
\end{aligned}
\end{equation}

\section{Overview}\label{sec:overview}

\hl{In the following subsections}, we define the structure of our traffic graph and outline the software architecture of our framework. 

\subsection{Traffic graph structure}\label{section:graph-representation}

Extending \pygeo's \texttt{HeteroData}\footnote{\heterodataurl}, \packagename's \ccode{CommonRoadData} class represents a heterogeneous traffic graph encapsulating nodes of both the vehicle ($\text{\fauxsc{v}}$) and lanelet ($\text{\fauxsc{l}}$) node type, as well as the structural metadata for the specific graph instance.
Formally, we have that ${\mathcal{A} = \{\text{\fauxsc{v}}, \text{\fauxsc{l}}\}}$ and ${\mathcal{R} = \{\text{\fauxsc{l2l}}, \text{\fauxsc{v2v}}, \text{\fauxsc{v2l}}, \text{\fauxsc{l2v}}\}}$.

\begin{figure*}[h]
    \centering
    \begin{overpic}[trim=250 100 200 200, clip, width=1.0\linewidth,tics=10]
   {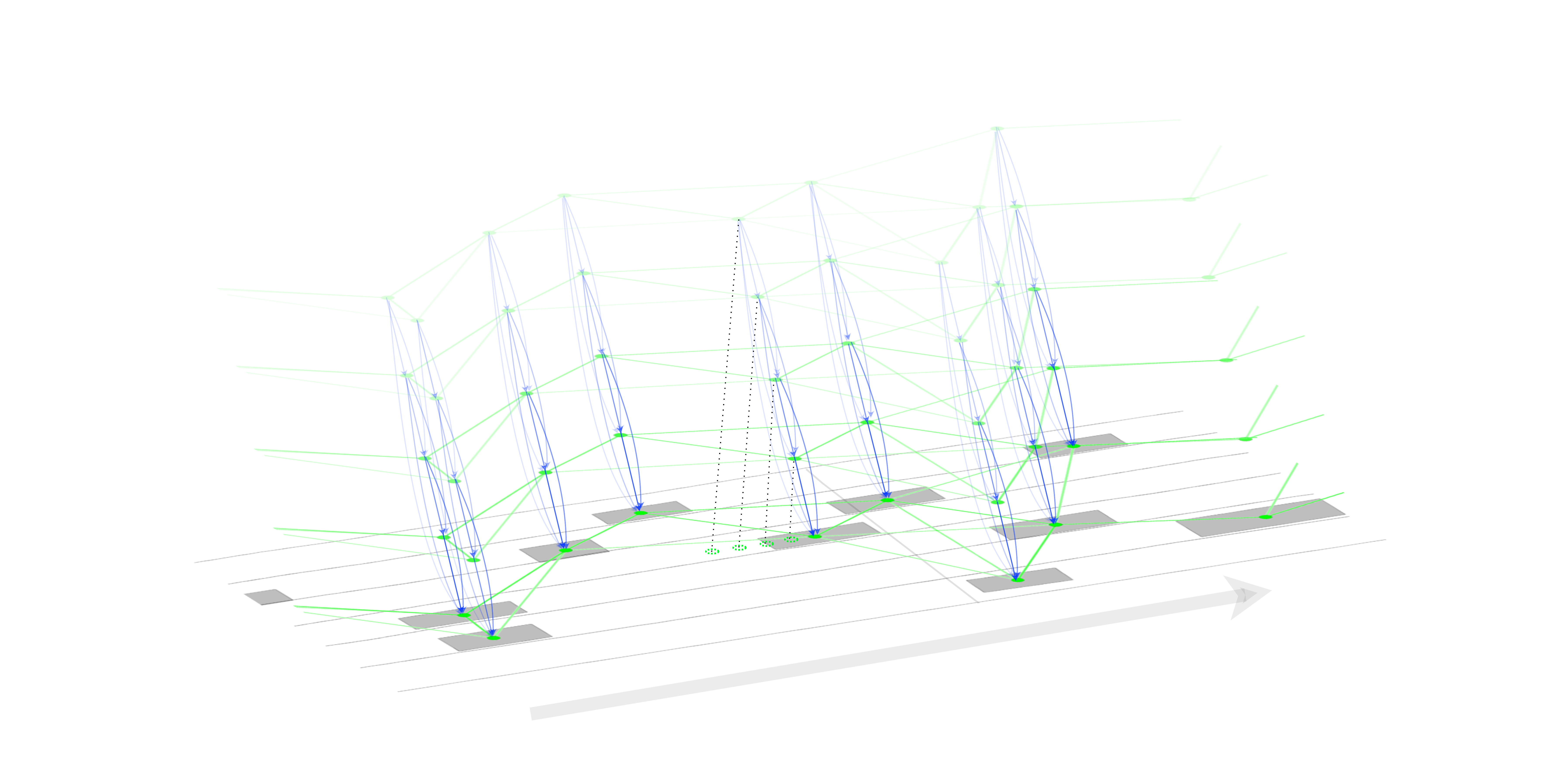}
    \caption{Visualization of Voronoi-based \textsc{v2v} and causal \textsc{vtv} edges for a highway \hl{scenario}. 
    Opaque colors denote the most recent edges while more transparent shades represent older ones.} 
 \put (41,10.5) {\footnotesize$t-4$}
 \put(46.2, 11.4){\rotatebox{9}{\ldots}}
 \put (51.2,11.9) {\footnotesize$t$}
 \put (1.8,8.5) {\footnotesize$t$}
 \put (52.7,1.0) {\rotatebox{9}{Direction of traffic}}
 \put (1.8,15.0) {\footnotesize$t-1$}
 \put (1.8,22.0) {\footnotesize$t-2$}
 \put (38,23.0) {\color{britishracinggreen}\textsc{v2v}}
 \put (29.0,25.0) {\rotatebox{-45}{\color{blue}\textsc{vtv}}}
 \put(0.5, 7.5){\color{black}\vector(0, 1){31}}
 \put (1.8,29.0) {\footnotesize$t-3$}
 \put (1.8,35.5) {\footnotesize$t-4$}\label{fig:vtv_edges}
   \end{overpic}
\end{figure*}
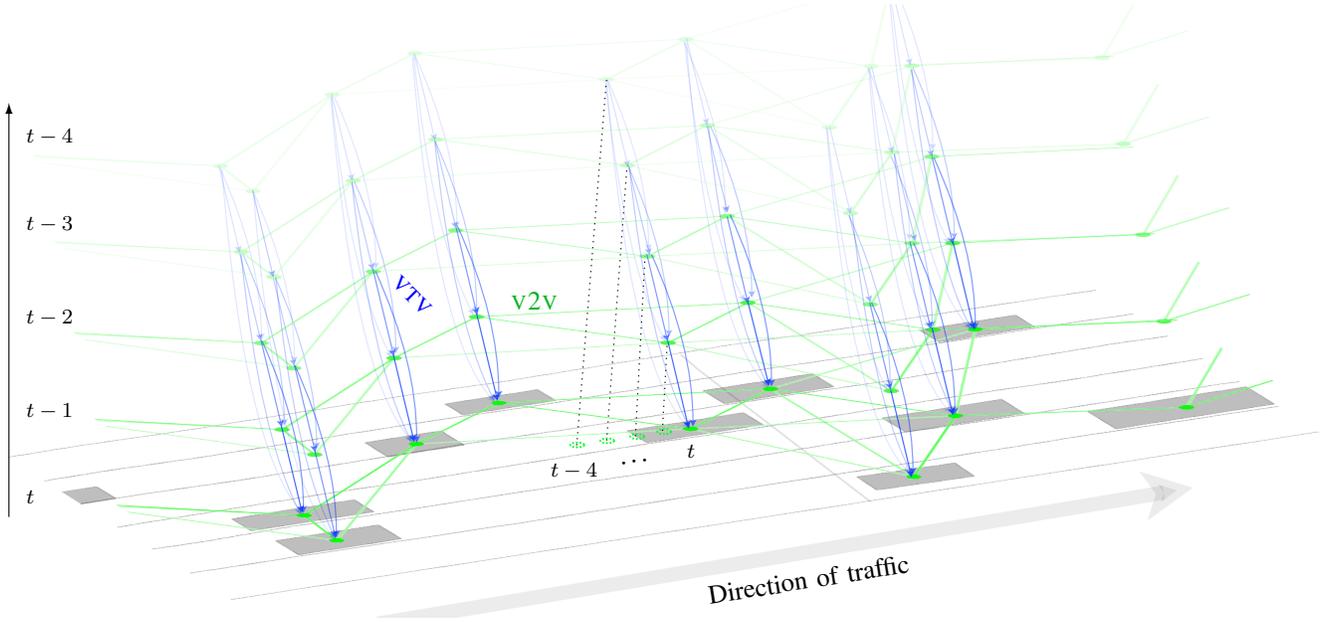

\hll{In order to capture temporal vehicle interactions with GNN-based message-passing schemes}, we further extend the \ccode{CommonRoadData} class \hl{by} the time dimension with \ccode{CommonRoadTemporalData}, where $\mathcal{R}$ is augmented by \hl{the temporal} \text{\fauxsc{vtv}} \hl{edge type}. The resulting \textit{temporal graph}~\cite{jain_2016_spatio_temporal_rnn}, \hll{as shown in} \cref{fig:vtv_edges}, \hl{intrinsically encodes the temporal dimension by unrolling the traffic graph over time. Vehicle nodes are repeated to capture vehicle states} \hll{at past timesteps}\hl{, and temporal edges encode the time difference between them.} 

The \hll{heterogeneous} structure of our graphical data representation is summarized in \cref{tab:commonroad-data-temporal-overview} and outlined in the following paragraphs:

\subsubsection{Lanelet nodes $(\mathcal{V}_L)$}\label{section:lanelet-nodes}
\hl{Lanelet nodes map to the lanelets in} $\mathcal{L}^{cr}$\hl{, with each lanelet} $L$ \hl{being represented as a graph node.}
\hl{The corresponding node features encode the geometric properties of the respective lanelets.}
Using the general notation $^{L}\square$, we denote the lanelet-local transformed polyline coordinates as ${}^{L}S_{L, l}$, ${}^{L}S_{L, c}$, and ${}^{L}S_{L, r}$. Here, the vertex coordinates are transformed to the lanelet-local coordinate frame according to its origin position ${\vect{p}_L = \vec{s}_{c,1}}$ and its orientation ${\theta_L = \atantwo{\left( \nicefrac{\vec{s}_{c,2}}{\vec{s}_{c,1}} \right)}}$.
In the following, we also let the function interpretations $\tilde{S}_{L, \square}(\cdot)$ and $\tilde{\theta}_L(\cdot)$ be defined via orthogonal projections onto the polylines, returning the position and orientation at a given arclength, respectively. 
%
%

\subsubsection{Lanelet-to-lanelet edges $(\mathcal{E}_{\textsc{l2l}})$}
\begin{figure}[h]
    \begin{subfigure}{1.0\textwidth}
        \centering
        \includegraphics[trim=62.6 135 50.5 135, clip, width=\textwidth]{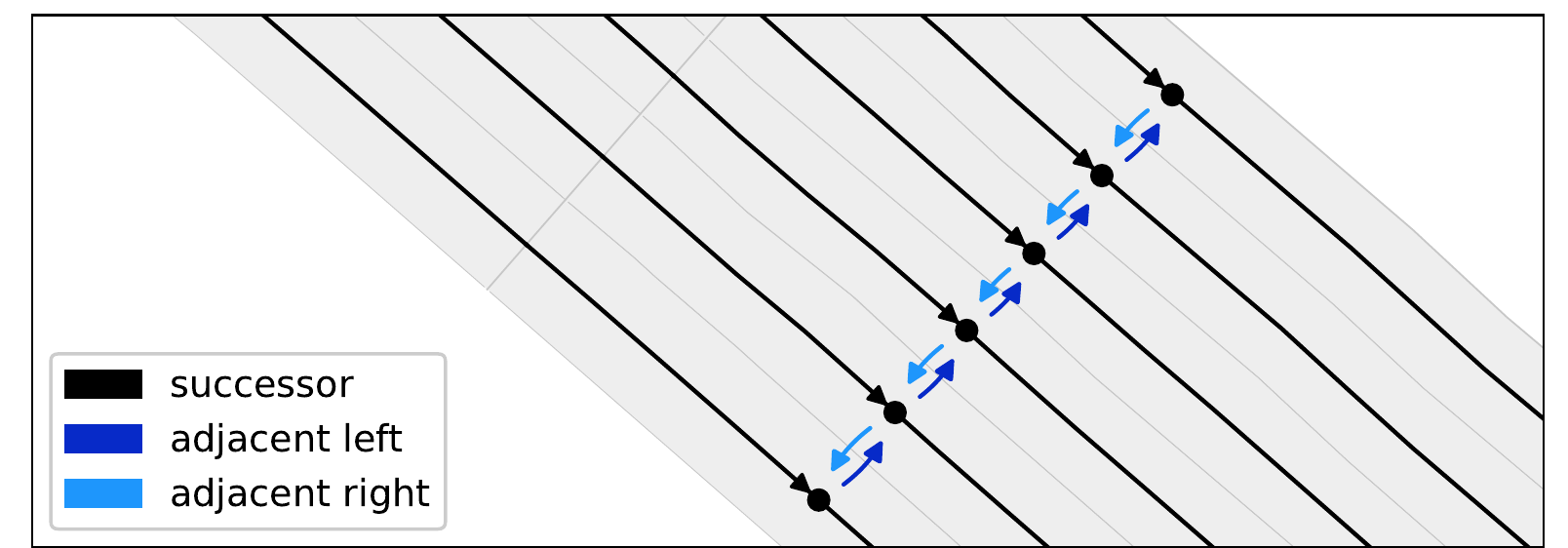}
        \caption{Highway scenario containing five parallel lanelets.\vspace{0.05cm}}
        \label{fig:subfig1}
    \end{subfigure}
    \begin{subfigure}{1.0\textwidth}
        \centering
        \includegraphics[trim=62.6 80 50.5 80, clip, width=\textwidth]{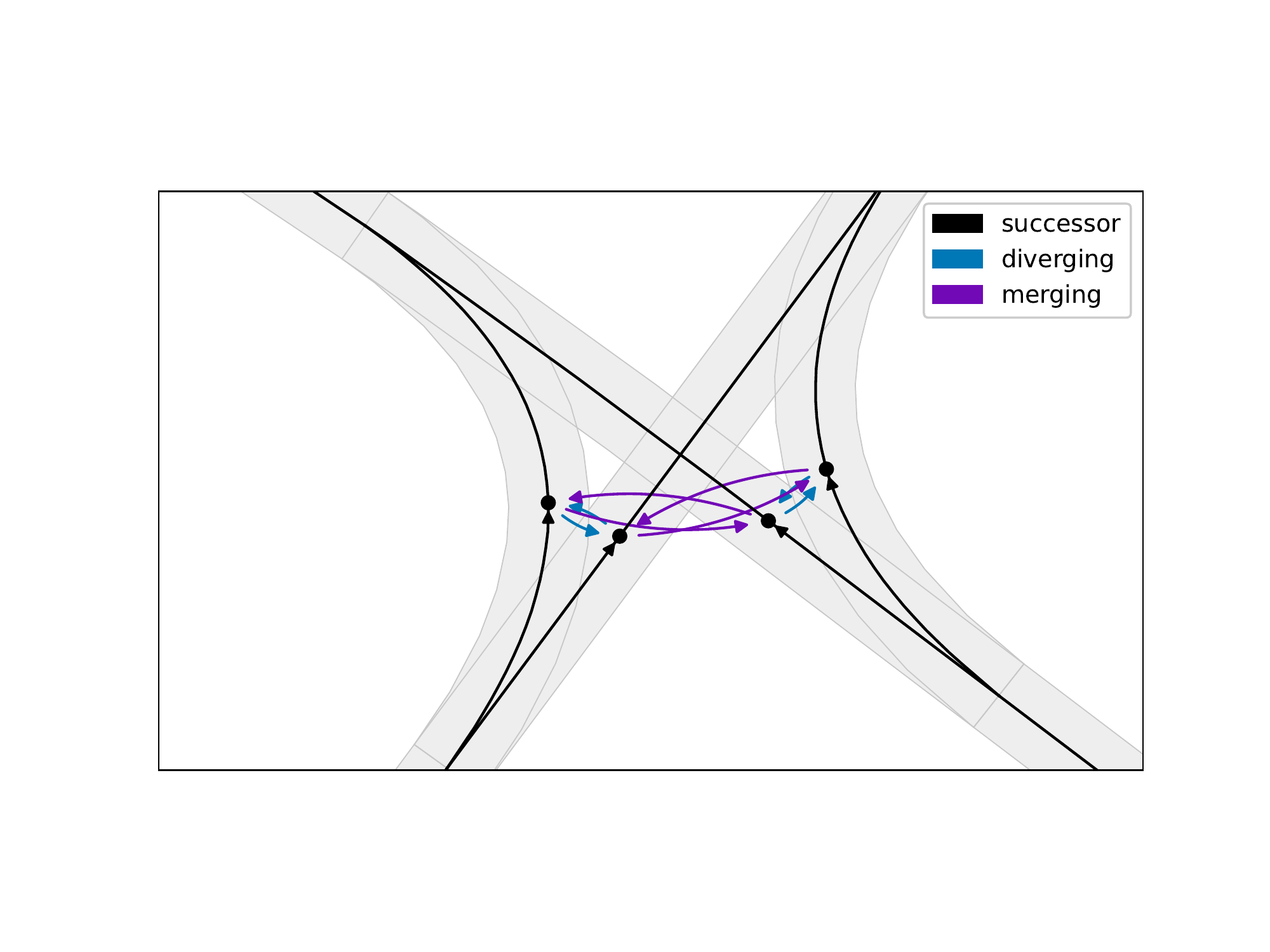}
        \caption{Intersection scenario highlighting merging and diverging lanelets.\vspace{0.05cm}}
        \label{fig:subfig2}
    \end{subfigure}
    \subfloat[Same road network as in (b), this time highlighting the conflicting lanelet nodes $L$ and $L^{'}$. The point of intersection is marked with a cross.\label{fig:l2l_intersection}]{%
        \centering
      \begin{overpic}[trim=62.8 100 50.2 100, clip, width=\textwidth]{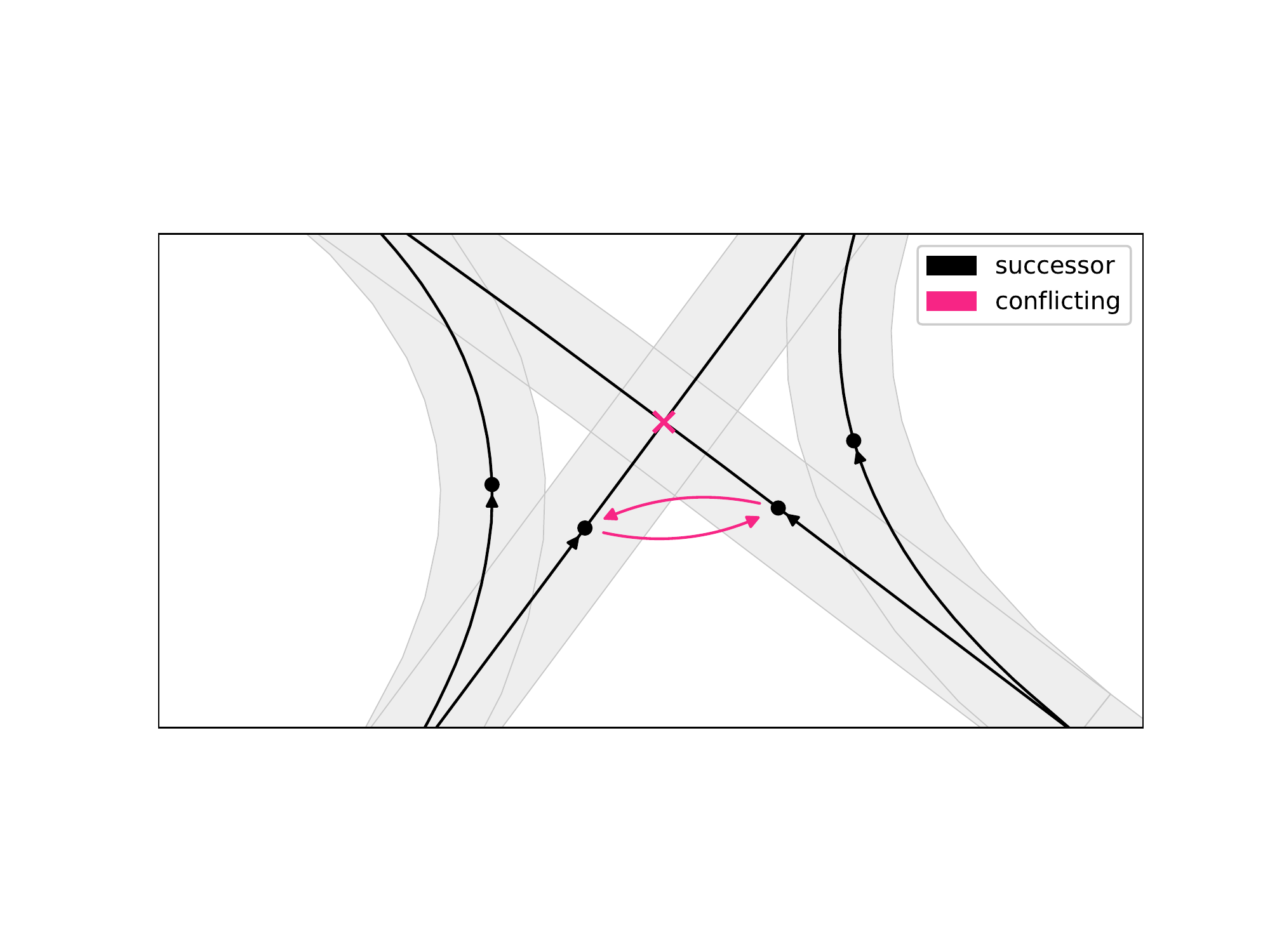}

 \put(54,30.3){\tikz \draw[dashed,->] (0,0)--(-2.1,0);}
 \put (79.6,29.6) {\footnotesize{$s_{L}$,  $s_{L^{'}}$}}
 \put (47,15) {\color{conflict}\scriptsize{ $L \rightarrow L^{'}$}}
 
 \put (60.3,24) {\footnotesize{ $L^{'}$}}
 \put (39.5,21.5) {\footnotesize{ $L$}}

      \end{overpic}
    }
    \caption{Lanelet graphs highlighting $\textsc{l2l}$ adjacency types.}
    \label{fig:lanelet_graph_conv}
\end{figure}
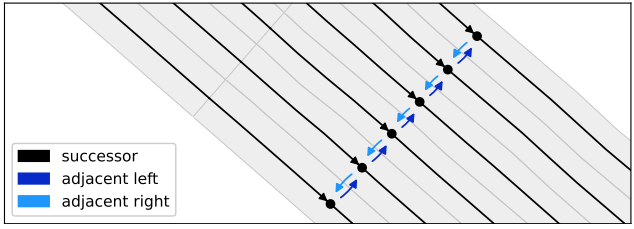
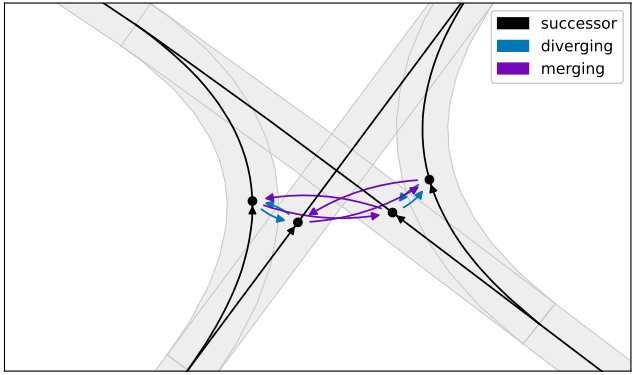
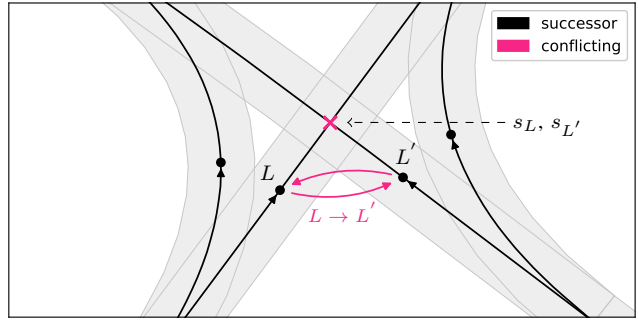

Lanelet-to-lanelet edges characterize the lanelet network topology and encode the spatial relationship between adjacent lanelets.
As illustrated by \cref{fig:lanelet_graph_conv},
we explicitly differentiate between the heterogeneous $\textsc{l2l}$ adjacency types $\mathcal{R}_{\textsc{l2l}}$ \hl{listed in} \cref{tab:l2lrelations} via the edge feature $\tau_{\mathcal{E}}$.
For an edge from $L$ to $L^{'}$, we also include their centerline arclength distance at the point of intersection as edge features, which we denote by $s_{L}$ and $s_{L^{'}}$. This is highlighted in~\cref{fig:l2l_intersection} for two conflicting lanelets.
%

\begin{table}[t]
\footnotesize
\centering
\caption{\textsc{l2l} adjacency types for edges $L \rightarrow L^{\prime}$. \hl{The considered adjacency types can be individually} \hll{selected} \hl{at the user's preference.}\vspace{-0.5cm}}
\begin{tabularx}{\linewidth}{cc}
    \multicolumn{2}{c}{} \\
    \toprule
    \textbf{Type} & \textbf{Interpretation}\\
    \midrule
		predecessor & $L$ continues the driving corridor of $L^{\prime}$ \\
		successor & $L^{\prime}$ continues the driving corridor of $L$ \\
		adjacent left & $L$ is left-adjacent to $L^{\prime}$ \\
		adjacent right & $L$ is right-adjacent to $L^{\prime}$ \\
		merging & $L$ and $L^{\prime}$ share a common successor (symmetric) \\
		diverging & $L$ and $L^{\prime}$ share a common predecessor (symmetric) \\
		conflicting & $L$ and $L^{\prime}$ cross each other (symmetric) \\
    \bottomrule
\end{tabularx}
    
    \label{tab:l2lrelations}
\end{table}
\subsubsection{Vehicle nodes $(\mathcal{V}_V)$}
Vehicle nodes are inserted according to the states of the currently present vehicles and identified by their CommonRoad IDs. \hlll{As shown in} \cref{fig:vtv_edges}, \ccode{CommonRoadTemporalData} additionally includes past vehicle states in the graph representation\hl{: here, a vehicle's state history is captured by time-attributed (but otherwise identical) vehicles nodes inserted at each timestep}.

\subsubsection{Vehicle-to-vehicle edges $(\mathcal{E}_{\textsc{v2v}})$}\label{section:vehicle-to-vehicle-edges}

\begin{figure}[h]
    \begin{subfigure}{1.0\textwidth}
        \centering
        \includegraphics[trim=0 530 0 530, clip, width=\textwidth]{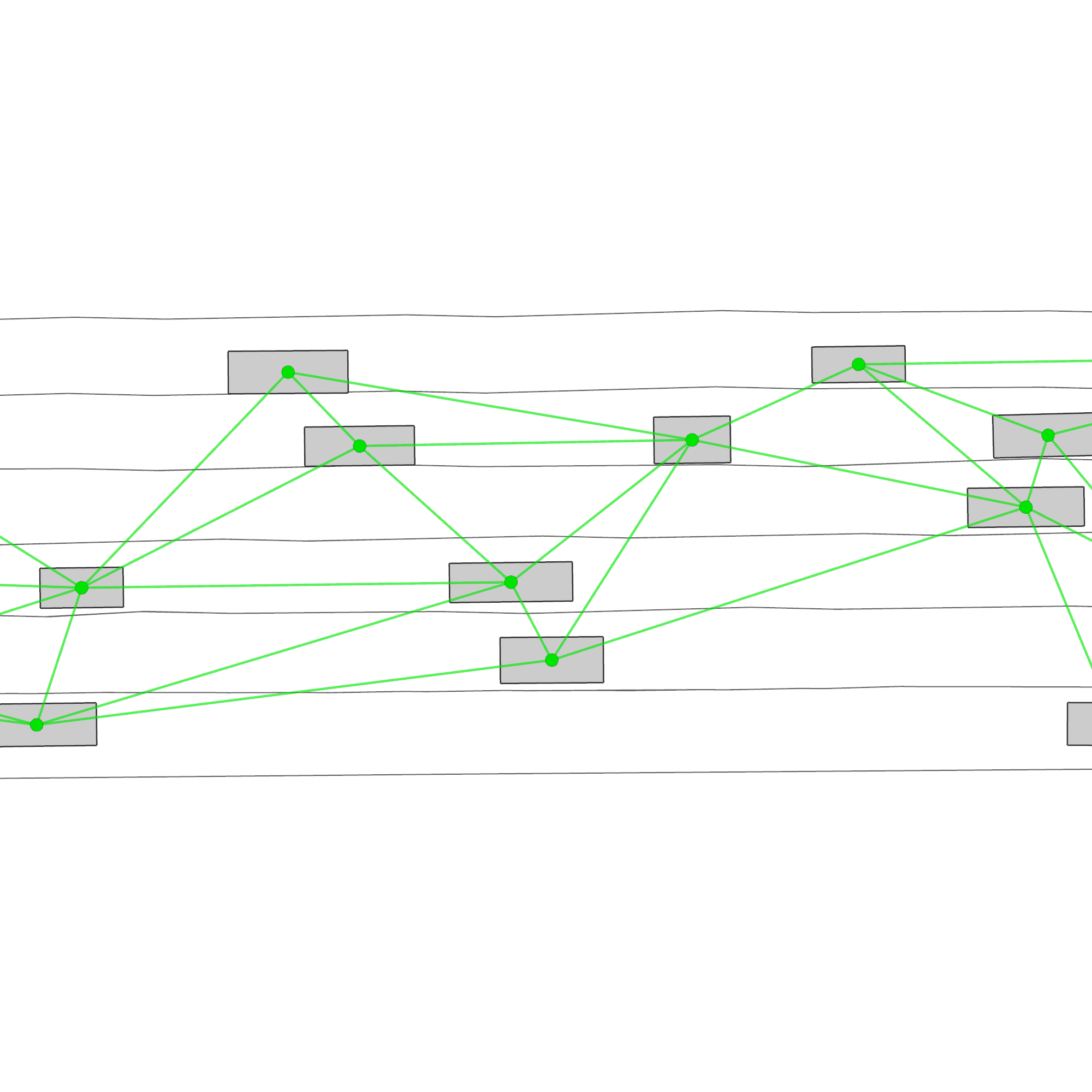}
        \caption{\code{VoronoiEdgeDrawer} inserts edges according to a Delaunay triangulation.}
        \label{fig:edge_drawers_demo_voro}
    \end{subfigure}
    \begin{subfigure}{1.0\textwidth}
        \centering
        \includegraphics[trim=0 530 0 550, clip, width=\textwidth]{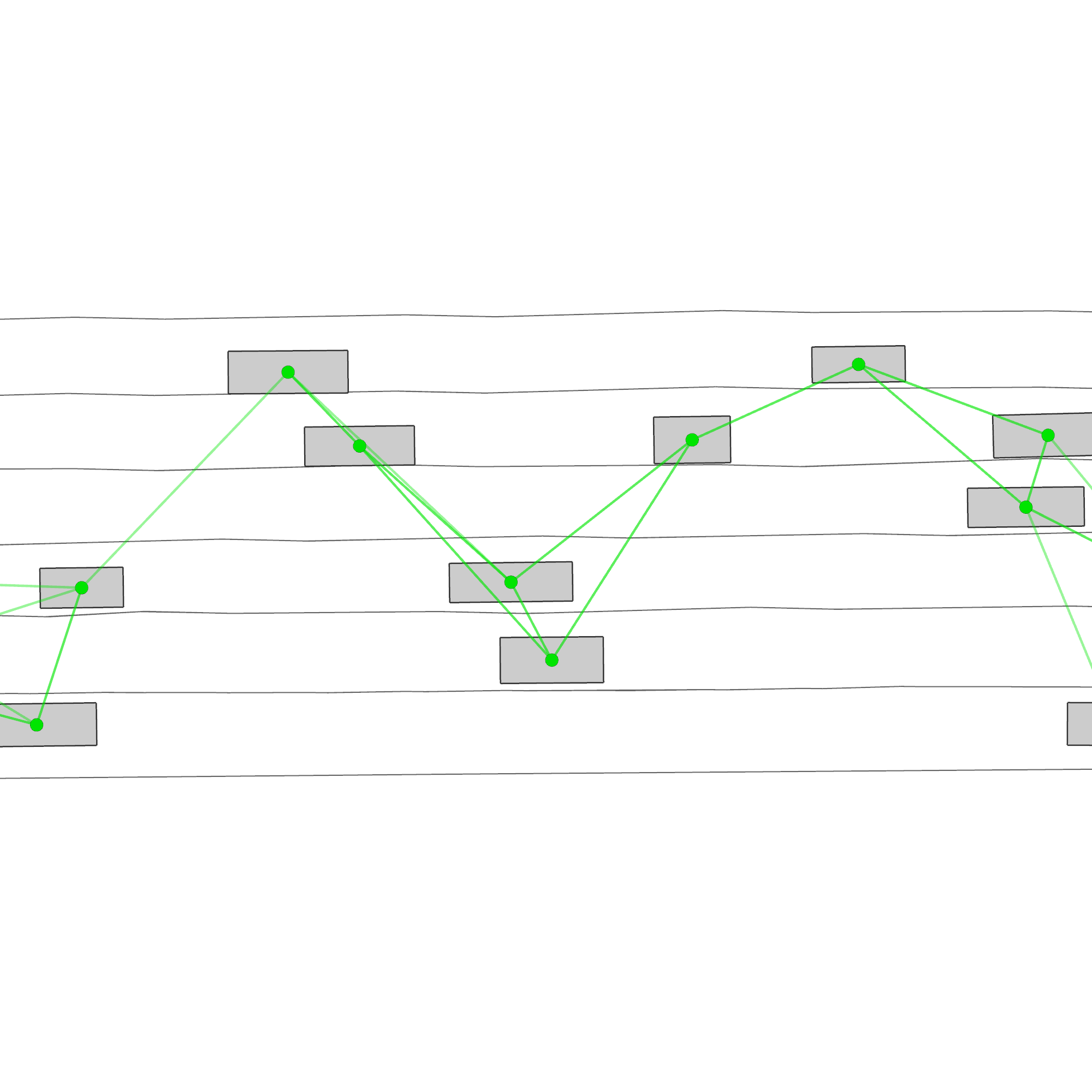}
        \caption{\code{KNearestEdgeDrawer} connects the $k$ closest vehicles (here with $k=3$).}
        \label{fig:subfig2}
    \end{subfigure}
    \caption{$\textsc{v2v}$ edges drawn by two different implementations.} 
    \label{fig:edge_drawers_demo}
\end{figure}

Vehicle-to-vehicle \hlll{edges} capture the interaction between vehicles at each timestep. \hl{The relative pose of connected vehicles is encoded as edge features according to the vehicle-local coordinate frame originating at the center of the source vehicle.} 
Users \hl{can} provide a specific implementation of our \text{\fauxsc{v2v}} \textit{edge drawer} protocol \hl{to encode which }\text{\fauxsc{v2v}} \hl{relations are relevant for their} \hll{task.}
Our framework offers \hllll{developers} a set of standard edge drawer implementations, two of which are depicted in \Cref{fig:edge_drawers_demo}.

\begin{table}[t]
\footnotesize
\centering
\caption{Overview of node ($\mathcal{X}_{\mathcal{V}}$) and edge ($\mathcal{X}_{\mathcal{E}}$) features.\vspace{-0.2cm}}
\renewcommand{\arraystretch}{1.275} 
\begin{tabularx}{\linewidth}{crllc}
\toprule
\textbf{Type}                   & \textbf{Feature}         & \textbf{Notation}                                 & \textbf{Unit}  & \textbf{Size} \\ \midrule

\multirow{6}{*}{$\underset{\underset{(L)}{}}{\textsc{l}}$}   & Position           & $\vect{p}_L$                                        & \SI{}{\meter} & 2 \\
                                & Length             & $\lVert L \rVert_2$                                   & \SI{}{\meter} & $1$  \\
                                & Orientation        & $\theta_L$                                          & \SI{}{rad} & $1$  \\
                                & Left vertices              & ${}^L V_l$                                          & \SI{}{\meter} & $2 \cdot N_{L}$ \\
                                & Right vertices             & ${}^L V_r$                                          & \SI{}{\meter} & $2 \cdot N_{L}$  \\
                                & Custom feature vector      & $\vect{x}_v^{(\textsc{l})}$                          & \NA  &            \\ \midrule

\multirow{8}{*}{$\underset{\underset{(V)}{}}{\textsc{v}}$}   & Position                   & $\vect{p}_V$                                        & \SI{}{\meter} & 2 \\
                                & Orientation                & $\theta_V$                                          & \SI{}{rad}  & $1$   \\
                                & Yaw-rate                & $\dot{\theta}_V$                                          & \SI{}{rad/\second} & $1$ \\
                                & Velocity                   & $\dot{\vect{p}}_V$                                        & \SI{}{\meter/\second} & $2$ \\
                                & Acceleration                   & $\vect{\ddot{p}}_V$     & \SI{}{\meter/\second^2} & $2$ \\
                                & Vehicle width                   & $w_V$ & \SI{}{\meter} & $1$ \\
                                & Vehicle length                   & $l_V$ & \SI{}{\meter} & $1$ \\

                                & Custom feature vector      & $\vect{x}_v^{(\textsc{v})}$                          & \NA &         \\ \midrule

\multirow{7}{*}{$\underset{\underset{(L \rightarrow L^{'})}{}}{\textsc{l2l}}$}
                                & Distance\textsuperscript{a}  & $\lVert \vect{p}_{L^{'}} - \vect{p}_L \rVert_2$                                          & \SI{}{\meter} & 1  \\
                                & Relative position\textsuperscript{a}                   & $^{L}\vect{p}_{L^{'}}$  & \SI{}{\meter} & $2$  \\
                                & Relative orientation\textsuperscript{a}                   & $\theta_{L^{'}} - \theta_L$  & \SI{}{rad} & $1$  \\

                                & Intersection (source) & $s_{L}$                                          & \SI{}{\meter} & $1$  \\
                                & Intersection (target) & $s_{L^{'}}$                                          & \SI{}{\meter} & $1$ \\
                                & Adjacency type                  & $\tau_{\mathcal{E}}$ & \NA & $1$               \\
                                & Custom feature vector      & $\vect{x}_e^{(\textsc{l2l})}$                       &  \NA &              \\ \midrule

\multirow{6}{*}{$\underset{\underset{(V \rightarrow V^{'})}{}}{\textsc{v2v}}$}
& Distance                   & $\lVert \vect{p}_{V^{'}} - \vect{p}_{V} \rVert_2$                                         & \SI{}{\meter} & 1 \\
 & Relative position      & $^{V}\vect{p}_{V^{'}}$  & \SI{}{\meter} & 2 \\

                                & Relative orientation                   & $\theta_{V^{'}} - \theta_{V}$                                        & \SI{}{rad} & 1 \\
                                & Relative velocity                   & $^{V}\dot{\vect{p}}_{V^{'}} - ^{V}\!\dot{\vect{p}}_{V}$                                         & \SI{}{\meter/\second} & $2$  \\
                                & Relative acceleration                   & $^{V}\ddot{\vect{p}}_{V^{'}} - ^{V}\!\ddot{\vect{p}}_{V}$                                         & \SI{}{\meter/\second^2} & $2$ \\
                           & Custom feature vector      & $\vect{x}_e^{(\textsc{v2v})}$ & \NA & \\ \midrule

\multirow{3}{*}{$\textsc{vtv}$\textsuperscript{b}}
& \textit{\fauxsc{v2v} features} & \ditto & \ditto & \ditto \\
& Time delta               & $\Delta t_{e}$                                      & \SI{}{\second} & $1$ \\
                                 & Custom feature vector      & $\vect{x}_e^{(\textsc{vtv})}$                        & \NA  &         \\ \midrule

\multirow{7}{*}{$\underset{\underset{(V \rightarrow L)}{}}{\textsc{v2l}}$}
& Left distance      & $d_{V, l}^{L}$  & \SI{}{\meter} & 1  \\
& Right distance      & $d_{V, r}^{L}$  & \SI{}{\meter} & 1  \\
& Lateral offset      & $d_{V, e}^{L}$  & \SI{}{\meter} & 1  \\
                                 & Heading error                   & $\theta_{V, e}^{L}$                                         & \SI{}{rad} & $1$ \\
                                & Projected arclength                   & $s_{V}^{L}$                                         & \SI{}{\meter} & $1$  \\
                                & Normalized arclength                   & $\nicefrac{s_{V}^{L}}{\lVert L \rVert_2}$                                         & \NA & $1$  \\
                                & Custom feature vector      & $\vect{x}_e^{(\textsc{v2l})}$                        &  \NA &  \\ 
\bottomrule
%

\end{tabularx} \par
\bigskip\vspace{-0.25cm}
\hspace{-1.42cm}\textsuperscript{a}\hl{measured between the respective lanelet coordinate frames.}\\
\hspace{-2.835cm}\textsuperscript{b}only relevant for \textit{CommonRoadTemporalData}.\vspace{-6pt}
\label{tab:commonroad-data-temporal-overview}
\end{table}
%

\subsubsection{Vehicle-to-lanelet edges $(\mathcal{E}_{\textsc{v2l}})$}

Vehicle-to-lanelet edges relate vehicles to the underlying road infrastructure. Our framework offers two assignment strategies for drawing the edges:

\begin{enumerate}
    \item \textbf{Center}: Each vehicle is connected to all lanelets that contain the vehicle center point. 
    \item \textbf{Shape}: Each vehicle is connected to all lanelets that \hl{intersect with the} vehicle shape. \hl{This constitutes a superset of the edges drawn by the \textit{Center} strategy.}
\end{enumerate}

The associated edge features describe the relative pose of the vehicle with respect to the curvilinear lanelet coordinate frame~\cite{hery_2017_curvilinear}.
For a given edge from vehicle $V$ to lanelet $L$, we denote the orthogonal distance from the lanelet's left and right boundaries to the vehicle center as
\begin{equation}
\begin{aligned}
d_{V, l}^{L} &= \lVert \tilde{S}_{L, l}(\vect{p}_{V}) - \vect{p}_{V} \rVert_2, \\
d_{V, r}^{L} &= \lVert \tilde{S}_{L, r}(\vect{p}_{V}) - \vect{p}_{V} \rVert_2.
\end{aligned}
\end{equation}
Further, we define the signed offset from the centerline as 
\begin{equation}
    d_{V, e}^{L} = \frac{d_{V, l}^{L} - d_{V, r}^{L}}{2}
\end{equation}
\hl{and} let $s_{V}^{L} \in [0, \lVert L \rVert_2]$ denote the centerline arclength from the lanelet origin $\vect{p}_{L}$ to $\tilde{S}_{L, c}(\vect{p}_{V})$.
\hl{Finally}, the orientation difference 
is \hlll{given} by 

\begin{equation}
    \theta_{V, e}^{L} = \tilde{\theta}_L(\vect{p}_{V}) - \theta_V.
\end{equation} 

\subsubsection{Vehicle-temporal-vehicle edges $(\mathcal{E}_{\textsc{vtv}})$}

\ccode{CommonRoadTemporalData} additionally contains temporal vehicle edges for encoding temporal dependencies in the traffic graph:
Letting $v_t$ and $v_{t^{'}}$ denote two vehicle nodes emerging from one vehicle instance, a temporal edge ${e\in \mathcal{E}_{\textsc{vtv}}}$ connects the vehicle object to itself at two different timesteps, as illustrated by \cref{fig:vtv_edges}. The temporal separation between the nodes is defined as the elapsed time ${\Delta t_e = t - t^{'}}$. 

%
\hl{As for $\textsc{v2v}$}, \packagename~lets users specify a \hl{custom} \code{temporal edge drawer} for defining the exact \text{\fauxsc{vtv}} graph structure. 
The default \ccode{CausalEdgeDrawer} inserts directed temporal edges between a historic vehicle node and its future realization at up to  $T_{max}^{\textsc{vtv}}$ future time steps. 
This constrains the flow of the \text{\fauxsc{vtv}} edges to be forward in time.

\subsection{Software architecture}

Next, we outline the principal components of our software architecture in a bottom-up approach, by detailing the \hl{pipeline for} collecting graph datasets from CommonRoad scenarios. An architecture overview is given by~\cref{fig:high-level}.

\subsubsection{Design principles}\label{sec:design-principles}

\begin{figure*}[t]
    \centering

    \begin{overpic}[width=1.0\linewidth]{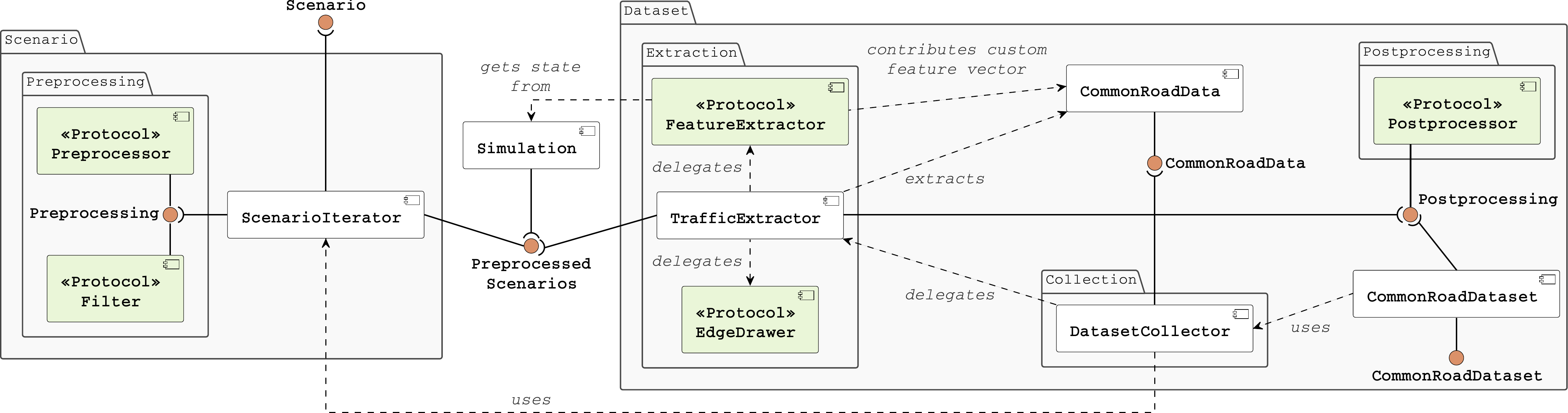}

   \caption{High-level software architecture for \packagename~shown as a UML 2 component diagram~\cite{Booch04_uml}.} 

 \label{fig:high-level}
   \end{overpic}
\end{figure*}

\begin{figure*}[H]
    \centering
\includegraphics[width=1.0\linewidth]{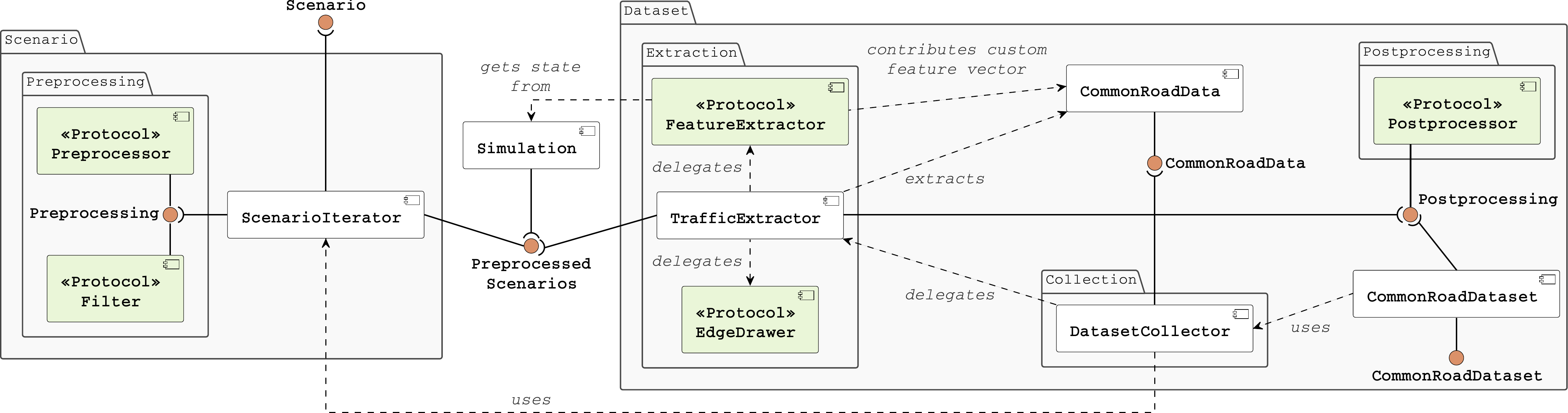}
    \caption{High-level software architecture for \packagename~shown as a UML 2 component diagram~\cite{Booch04_uml}.}
    \label{fig:high-level}
\end{figure*}

Conforming to the \textit{single responsibility principle}~\cite{Martin17_clean_software}, \packagename~is composed of distinct modules with clearly defined roles. To encapsulate volatile processing routines that may require frequent modifications, the \textit{strategy pattern}~\cite{gamma1994design} is employed. This lets the user realize specific behaviors through composition, making it possible to extend the framework to satisfy different requirements and use cases. Users can introduce or substitute implementations without interfering with the rest of the framework, avoiding unintended side effects and minimizing debugging, testing, maintenance, and refactoring efforts.

Next, we introduce the core functionalities of our framework on the basis of these design principles. \cref{tab:extendable_components} \hl{summarizes the options available to customize graph extraction:}

\begin{table}[H]
    \footnotesize
    \begin{tabularx}{\linewidth}{ llcc }
    \multicolumn{4}{c}{} \\
    \toprule
    \textbf{Component} & \textbf{Target} & \textbf{Count} & \textbf{Default}\\
    \midrule
		\textit{Preprocessors} & input scenario  & \textit{many} & $\emptyset$  \\
		\textit{Filters} & input scenario & \textit{many} & $\emptyset$    \\
		\textsc{v2v} \textit{edge drawer} & $\mathcal{E}_{\fauxsc{v2v}}$ & \textit{one} & \footnotesize{\texttt{VoronoiEdgeDrawer}}  \\
		\textsc{vtv} \textit{edge drawer} &  $\mathcal{E}_{\fauxsc{vtv}}$ & \textit{one} & \footnotesize{\texttt{CausalEdgeDrawer}}   \\
		\textit{Feature extractors} & $\mathcal{X}_{\mathcal{V}}$,  $\mathcal{X}_{\mathcal{E}}$ & \textit{many}  & $\emptyset$   \\
		\textit{Postprocessors} & $\mathcal{G}$ & \textit{many} & $\emptyset$   \\
    \bottomrule
    \end{tabularx}
    \caption{\hl{Summary of} \packagename\hl{'s base protocols facilitating user-specified behaviors.}\vspace{-0.5cm}}
    \label{tab:extendable_components}
\end{table}

\subsubsection{Scenario preprocessing and filtering}
Our framework supports arbitrary preprocessing and filtering of the input scenarios to ensure that they satisfy the user requirements.
As an example, \ccode{TrafficFilter(min=10)} lets users exclude scenarios with less than $10$ vehicles from the collected graph dataset. Further, suppose that we want a higher-fidelity view of the lanelet network, where no lanelet exceeds $20$ meters in length: this can be achieved by using \packagename's built-in \ccode{SegmentLanelets} \textit{preprocessor}, which results in a higher lanelet node density. Using our composition syntax that realizes arbitrary chaining and grouping of preprocessing and filtering operations, these behaviors can be effortlessly combined via \\ \ccode{TrafficFilter(min=10) >> SegmentLanelets(size=20)}.

\subsubsection{Feature extraction}
The extraction of the graph features, i.e., $\mathcal{X}_{\mathcal{V}}$ and $\mathcal{X}_{\mathcal{E}}$, 
is carried out through \code{feature extractors} operating on the scenario objects. The feature extractors receive a \textit{simulation} object, \hll{which provides access} to inherent and derived state information of the scenario at the current timestep. \hl{By maintaining an internal state,} \hll{feature extractors} \hl{also support time-dependent features.}

In our framework, we distinguish between node- and edge-level feature extractors, with the latter operating on pairs of entities according to $\mathcal{E}$. 
\hll{In accordance with} \packagename\hl{'s guiding design principles, users can devise their own implementation to augment the extracted graphs by arbitrary custom features without modifying the} \packagename~\hl{source code.}
To reduce the initial setup time for new \hll{users}, \packagename~\hl{also} offers users a selection of pre-configured feature extractor implementations designed to meet the most common needs and requirements.
\subsubsection{Postprocessors}
In contrast to the aforementioned node- and edge-level feature extractors, \code{postprocessors} operate directly on the graph instances in a deferred manner.
As such, they \hll{are intended to supplement the flexibility we concede in the otherwise rigorous graph extraction procedure. User-devised} \hl{postprocessing procedures} can serve numerous practical purposes, \hll{e.g.,}
\begin{itemize}
    \item to complement the functionality of feature extractors by facilitating the computation of \hl{graph-global} features, e.g., an indicator for whether the current traffic scene corresponds to a traffic jam or not;
    \item to modify the graph structure, e.g., by the removal or insertion of nodes and edges.
\end{itemize}

\hl{The postprocessors are executed after the initial graph extraction, but prior to dataset generation. Furthermore, they can also be applied while loading an existing dataset.}

\subsubsection{Traffic graph extraction}
The execution of edge drawing, feature extraction and postprocessing is orchestrated by a \code{traffic extractor},
\hl{which} manages the creation of a \ccode{CommonRoadData} instance at a single timestep $t$.
It also handles the declaration of metadata attributes such as node IDs.

The traffic extractor class is complemented by the \code{temporal traffic extractor} for the purpose of extracting temporal graph representations. Internally relying on a \hll{regular} traffic extractor, it maintains a cache of the $n$ preceding \hll{ordinary} \ccode{CommonRoadData} instances at all times. When called upon, it returns the \ccode{CommonRoadTemporalData} at the current timestep by merging the cached graph sequence into a single entity. 
\hl{The} temporal extractor delegates the \hll{temporal edge} construction to the provided \code{temporal edge drawer} and the computation of custom temporal features to the \text{\fauxsc{vtv}} feature extractors provided by the user. The regular and temporal extraction procedures are summarized by~\cref{alg:data-creation} and~\cref{alg:temp-data-creation}, respectively.

\begin{algorithm}
\captionsetup{font=footnotesize} 
\begin{minipage}{\dimexpr\linewidth-\SmallMargin\relax}
  \caption{Extraction of \code{CommonRoadData} by \code{TrafficExtractor}.}\label{alg:data-creation}
\footnotesize
  \begin{algorithmic}[1]
\Configuration
    \State Scenario \code{scenario}
    \State Edge drawer $D^{\fauxsc{v2v}}$
    \State Set of feature extractors $\mathcal{F}$
    \State Sequence of postprocessors $\mathcal{P}_{post}$
\Initialization
    \State \code{simulation} $\leftarrow$ \code{Simulation}$ ($\code{scenario}$)$
\Require
    \State Timestep $t$
\Ensure
    \State \code{CommonRoadData}
\Proc \textsc{Extract}
    \State \code{state} $\leftarrow$ \code{simulation}$(t)$ \Comment{Current traffic state}
    \State \code{v2v\_edges} $\leftarrow$ $D^{\fauxsc{v2v}}$$($\code{state}$)$
    \State \code{features} $\leftarrow$ $\emptyset$ \Comment{Container for \fauxsc{v2v}, \fauxsc{l2l}, \fauxsc{v2l}, and \fauxsc{l2v} features}
    \For{\textbf{each} $f \in \mathcal{F}$} \hspace*{7.7em}\tikzmark{right} \tikzmark{top}
        \State \code{feature} $\leftarrow$ $f($\code{state}, \code{v2v\_edges}$)$
        \State \code{features} $\leftarrow$ \code{features} $\cup \; \code{feature}$
    \EndFor \tikzmark{bottom}
    \State \code{data} $\leftarrow$ \code{CommonRoadData}$($\code{state}, \code{v2v\_edges}, \code{features}$)$
    \For{\textbf{each} $p \in$ $\mathcal{P}_{post}$}  \hspace*{1em}\tikzmark{rightp} \tikzmark{topp}
        \State \code{data} $\leftarrow$ $p($\code{data}$)$
    \EndFor \tikzmark{bottomp}
    \Return \code{data}
    \vspace{-0.25cm}
\end{algorithmic}%
\AddNote{top}{bottom}{right}{Feature extraction}
\AddNote{topp}{bottomp}{rightp}{Postprocessing}
\end{minipage}%
\end{algorithm}

\begin{algorithm}
\captionsetup{font=footnotesize} 
 \caption{Extraction of \code{CommonRoadTemporalData}.}
 \label{alg:temp-data-creation}
\footnotesize
  \begin{algorithmic}[1]


\Configuration
    \State Data cache size $n$
    \State Single timestep traffic extractor $E_{regular}$\Comment{Ref.~\cref{alg:data-creation}}  
    \State Temporal edge drawer $D^{\fauxsc{vtv}}$
    \State Set of temporal feature extractors $\mathcal{F}^{\fauxsc{vtv}}$
    \State Sequence of temporal data postprocessors $\mathcal{P}_{post}^{\fauxsc{vtv}}$
\Initialization
    \State Empty data sample cache $\mathcal{C} \leftarrow$ \textsc{List}$($\code{max\_size=n}$)$
\Require
\State Timestep $t$
\Ensure
    \State \code{CommonRoadTemporalData}
\Proc \textsc{Extract}
    \State \code{data} $\leftarrow$ $E_{regular}$$(t)$
    \State $\mathcal{C}$.\code{put}$($\code{data}$)$
    \State \code{vtv\_edges} $\leftarrow$ $D^{\fauxsc{vtv}}$$(\mathcal{C})$
    \State \code{vtv\_features} $\leftarrow$ $\emptyset$ \Comment{Container for \fauxsc{vtv} features}
    \For{\textbf{each} $ f \in $ $\mathcal{F}^{\fauxsc{vtv}}$} \hspace*{9em}\tikzmark{right} \tikzmark{top}
        \State \code{feature} $\leftarrow$ $f(\mathcal{C}$, \code{vtv\_edges}$)$
         \State \code{vtv\_features} $\leftarrow$ \code{vtv\_features} $\cup \; \code{feature}$
    \EndFor \tikzmark{bottom}
    \State \code{t\_data} $\leftarrow$ \code{CommonRoadTemporalData}$(\mathcal{C}$, \code{vtv\_edges},  \code{vtv\_features}$)$
    \For{\textbf{each} $p \in$ $\mathcal{P}_{post}^{\fauxsc{vtv}}$}  \hspace*{1em}\tikzmark{rightp} \tikzmark{topp}
        \State \code{t\_data} $\leftarrow$ $p($\code{t\_data}$)$
    \EndFor \tikzmark{bottomp}
    \Return \code{t\_data}\textsuperscript{t}
    \vspace{-0.2cm}
  \end{algorithmic}
\AddNote{top}{bottom}{right}{Feature extraction}
\AddNote{topp}{bottomp}{rightp}{Postprocessing}
\end{algorithm}

\subsubsection{Dataset creation}
Based on the provided \hlll{traffic graph} extractor, the \code{dataset collector} offers an interface for generating a chronological sequence of graph instances from a specified scenario.
The collector iterates over consecutive timesteps until it reaches the end of the scenario lifetime, dynamically returning the extracted graph \hlll{objects}.
Whereas the static simulation mode corresponds to a replay of the recorded vehicle trajectories contained within the CommonRoad scenario, the interactive mode leverages an interactive traffic simulator for on-the-fly generation of realistic vehicle behavior. One such implementation is provided by the traffic simulation tool SUMO~\cite{lopez_microscopic_2018}, which we access via our CommonRoad interface~\cite{crsumo}.

Finally, the creation and accessing of a persistent graph dataset collected from a set of input scenarios is facilitated by the \ccode{CommonRoadDataset}, a full-fledged extension of the powerful \texttt{Dataset}\footnote{\footnotesize{\url{https://pytorch-geometric.readthedocs.io/en/latest/generated/torch_geometric.data.Dataset.html}}} class natively offered by \pygeo. As such, our dataset class allows users to easily perform common data operations such as batching, sampling, and parallelization \hl{on the collected dataset} during \hl{model} training. 
As the individual graph instances inherit from the base data representation used by \pygeo, \packagename~practitioners can \hlll{effortlessly} adopt their wide selection\footnote{\footnotesize{\url{https://pytorch-geometric.readthedocs.io/en/latest/modules/}}} of state-of-the-art GNN architectures. 
The dataset creation procedure is summarized by~\cref{alg:dataset-creation}.
\algnewcommand{\algorithmicgoto}{\textbf{go to}}%
\algnewcommand{\Goto}[1]{\algorithmicgoto~\ref{#1}}%

\begin{algorithm}[H]
\captionsetup{font=footnotesize} 
\caption{Creation of \code{CommonRoadDataset} from \code{CommonRoadData}.}\label{alg:dataset-creation}
\footnotesize
\begin{algorithmic}
\Configuration
    \State Composed preprocessor $P_{pre}$ \Comment{Chain of \code{preprocessors} and \code{filters}}
    \State Edge drawers $\mathcal{D}$ \Comment{Contains $D^{\fauxsc{v2v}}$ and (for temporal datasets) $D^{\fauxsc{vtv}}$}
    \State Set of feature extractors $\mathcal{F}$
    \State Sequence of postprocessors $\mathcal{P}_{post}$
\Require
    \State Set of scenarios $\mathcal{S}$ \Comment{CommonRoad scenario directory}
    \State Timesteps $T$ per scenario
\Ensure
    \State \code{CommonRoadDataset} 
\Proc \textsc{Create}
    \State \code{dataset} $\leftarrow$ \code{CommonRoadDataset}$()$
    \For{\textbf{each} \code{scenario} $\in$ \code{ScenarioIterator}$(\mathcal{S}, P_{pre})$}

        \State \code{simulation} $\leftarrow$ \code{Simulation}$ ($\code{scenario}$)$ \hspace*{2em} \tikzmark{top}
        \State $E$ $\leftarrow$ \code{Extractor}$($\code{simulation}, $\mathcal{D}$, $\mathcal{F}$, $\mathcal{P}_{post}$$)$ \tikzmark{right}
        \State \code{samples} $\leftarrow$ $\emptyset$
        \For{\code{t} $\leftarrow$ $1,2,\ldots,T$}
            \State \code{data} $\leftarrow$ $E$$($\code{t}$)$
            \State \code{samples} $\leftarrow$ \code{samples} $\cup \; \code{data}$
        \EndFor \hspace*{2em}\tikzmark{bottom}
        \State \code{dataset} $\leftarrow$ \code{dataset} $\cup \; \code{samples}$
    \EndFor
    \Return \code{dataset}
\end{algorithmic}
\AddNote{top}{bottom}{right}{Delegated to \code{DatasetCollector}}
\end{algorithm}
\section{Dataset and Experiment}

To demonstrate the usage of our framework, we published\footnote{ \footnotesize\url{https://commonroad.in.tum.de/datasets}} a graph-converted dataset with a diverse set of real-world road geometries. With the help of the dataset converter\footnote{{\footnotesize\url{https://commonroad.in.tum.de/tools/dataset-converters}}}, we use the NuPlan~\cite{caesar_nuplan_2022} dataset as our data source, \hll{due to the diversity in the incorporated locations and environments}. The extracted graph dataset contains \packagename's \hll{default} graph features as previously listed in~\cref{tab:commonroad-data-temporal-overview}.

\subsection{Experiment}

As an example use case of our framework, we briefly\footnote{The model implementation can be found in the \packagename~repository.} introduce a spatiotemporal \hl{GNN} model \hll{for vehicle trajectory prediction based on our} \ccode{CommonRoadTemporalData} \hlll{environment} \hll{representation}.
%
%
We implement an end-to-end trainable encoder-decoder architecture based on \hl{the message passing framework offered by} \pygeo:

\subsubsection{Encoder component}
The GNN encoder, for which we use an adapted version of the Heterogeneous Graph Transformer (HGT) architecture proposed in~\cite{hu_2020_hgt}, computes a node-level embedding for each vehicle via message passing-based aggregation. This fixed-sized vector encoding summarizes the social and map-related information required to predict its future movement in the current traffic environment.
In order to exploit all properties of the graph-structured data, we implement an edge-enhanced HGT architecture that computes attention weights and messages based on both node and edge features. To accelerate learning, we further carry out three (trainable) \hl{encoding} steps that are applied before the graph convolution layers:
\begin{itemize}
    \item For encoding the delta-time attribute $\Delta t_e$ of the \textsc{vtv} edges, we adopt Time2Vec~\cite{diniz_2022_time2vec}, a learnable vector representation of time that lets us capture periodic and non-periodic time series patterns.
    \item To get fixed-sized node representations for the lanelet geometries, we use a Gated Recurrent Unit (GRU)~\cite{cho2014properties} \hll{which} encodes the variable-length waypoint sequences ${}^L V_l$ and ${}^L V_r$.
    \item Finally, we adopt a learnable vector embedding to encode the \textsc{l2l} adjacency type $\tau_{\mathcal{E}}$.
\end{itemize}

\subsubsection{Decoder component}
\begin{figure}[h]
    \centering
    \fbox{\includegraphics[trim=1600 1400 1400 2600, clip, width=0.95\linewidth]{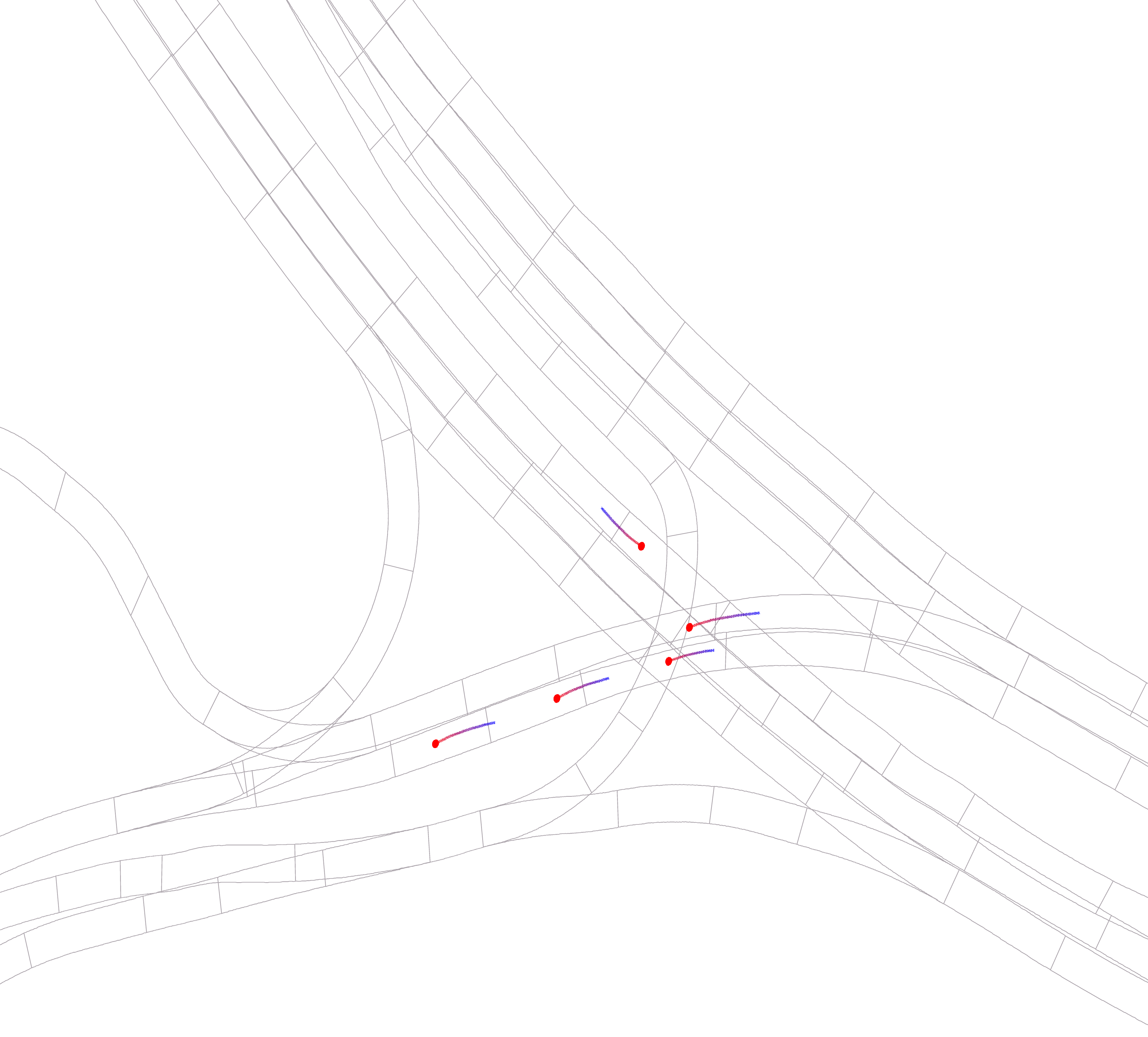} }
    \caption{Predicted trajectories from NuPlan's Singapore dataset.}
    \label{fig:predicted_trajectories}
\end{figure}

Based on the encoded vehicle representations, a GRU decoder network generates a fixed-length sequence of local position and orientation deltas for each vehicle, which is aggregated to obtain a sequence of predicted vehicles states. This is done by repeatedly updating the decoded vehicle state by a local transition in the coordinate frame of the previous state. The model is trained to minimize the average displacement error (ADE) between the predicted and ground-truth vehicle trajectories. An example of the resulting predictions with a prediction horizon of $1.0$~\SI{}{\second} and a time interval of $0.2$~\SI{}{\second} is shown in \Cref{fig:predicted_trajectories}, whereas the quantitative results are summarized in~\Cref{tab:results}.

\begin{table}[h]
    \footnotesize
    \begin{tabularx}{\linewidth}{ lccc }
    \multicolumn{4}{c}{} \\
    \toprule
    \textbf{Dataset split} & \textbf{Number of scenarios} & \textbf{ADE} [m] & \textbf{FDE} [m] \\
    \midrule
		\textit{Singapore} & 2372 & 0.106 & 0.227 \\
		\textit{Boston} & 938 & 0.138 & 0.316 \\
            \textit{Pittsburgh} & 1560 & 0.215 & 0.454 \\

    \bottomrule
    \end{tabularx}
    \caption{Experimental results from the cities included in the Nuplan dataset. Each experiment is trained and validated on the datasets collected in the same city. We use average displacement error and final displacement error (FDE) as evaluation metrics.}
\label{tab:results}
\end{table}
\section{Conclusion}\label{sec:conclusion}

This paper presents \packagename, an open-source Python package offering a standardized interface for map-aware graph extraction from traffic scenarios.
As a pioneering effort, it serves as a flexible framework that lets users collect custom graphical PyTorch datasets tailored to their research needs. Exemplified by our trajectory prediction implementation, it minimizes the time spent by researchers on writing boilerplate code for dataset collection. \packagename~\hl{complements the CommonRoad software platform}, which offers a converter tool for popular traffic datasets.
With ease of use and extension being our core design goals, \packagename's flexible interface is achieved by delegating and encapsulating all steps of the traffic graph creation. We invite researchers to contribute to \packagename~to further enhance its capabilities.

\section*{Acknowledgements}
This research was funded by the German Research Foundation grant AL 1185/7-1 and the Federal Ministry for Digital and Transport through the project KoSi.

\interfootnotelinepenalty=10000
\AtNextBibliography{\footnotesize}
\balance
\printbibliography

\end{document}